\documentclass[a4paper,fleqn]{article}
\usepackage[english]{babel}
\usepackage[authoryear]{natbib}
\usepackage{graphicx} 
\usepackage{float}
\usepackage{algorithm}  
\usepackage{algpseudocode}
\usepackage{color}
\usepackage{setspace}
\usepackage{amsmath}
\usepackage[squaren, Gray, cdot]{SIunits}
\usepackage{subcaption}
\usepackage{hyperref}
\usepackage{authblk}
\usepackage[a4paper, left=2cm, right=2.5cm, top=2.5cm, bottom=3.5cm]{geometry}

\title{\texttt{PyAWD}: A Library for Generating Large Synthetic Datasets of Acoustic Wave Propagation}

\author[1]{Pascal Tribel\footnote{ORCID: \href{https://orcid.org/0009-0006-3701-1223}{0009-0006-3701-1223}}}
\author[1]{Gianluca Bontempi\footnote{ORCID: \href{https://orcid.org/0000-0001-8621-316X}{0000-0001-8621-316X}}}

\affil[1]{Machine Learning Group, Université Libre de Bruxelles, Belgium \\ \url{http://mlg.ulb.ac.be/}}

\date{}

\begin{document}
\maketitle

\begin{abstract}
    Seismic data is often sparse and unevenly distributed due to the high costs and logistical challenges associated with deploying physical seismometers, limiting the application of Machine Learning (ML) in earthquake analysis. While simulation methods exist, no tool allows the generation of large datasets containing simulated measurements of the ground motion. To address this gap, we introduce \texttt{PyAWD}, a Python library designed to generate high-resolution synthetic datasets simulating spatio-temporal acoustic wave propagation in both two-dimensional and three-dimensional heterogeneous media. By allowing fine control over parameters such as the wave speed, external forces, spatial and temporal discretization, and media composition, \texttt{PyAWD} enables the creation of ML-scale datasets that capture the complexity of seismic wave behavior. We illustrate the library's potential with an epicenter retrieval task, showcasing its suitability for designing complex, accurate seismic problems that require advanced ML approaches in the absence or lack of dense real-world data. We also show the usefulness of our tool to tackle the problem of data budgeting in the framework of epicenter retrieval.
\end{abstract}


\section{Introduction}
        Earthquakes, as sudden and violent geological events, impact lives, infrastructures, and the environment on a large scale. Forecasting them and extracting precise information from measurements is challenging due to the complex interplay of seismic waves in heterogeneous propagation fields. Mapping problems relying on seismic data present an interesting case of multivariate time-series analysis where complex patterns occurring in the series can give valuable insights on certain tackled problems, such as earthquake epicenter retrieval, prediction of the next wave measurements, analysis of the moment of occurrence of events, …
        Machine Learning methods have shown promising results when applied to multivariate series analysis (\cite{cnn_location, eqconvmixer, phase-picking-cnn}). Seismic data is not an exception, and multiple regression and classification problems have been tackled by ML approaches applied to such spatio-temporal series (\cite{dl_epicenter, cnn_location, phase-picking-attention, structure-classification}). 
        Currently, ground motion is typically recorded using devices called \textit{seismometers}. Such a device continuously measures ground speed and acceleration in the three spatial dimensions. It is expensive and cannot cover a wide area, leading to sparse datasets. Traditional seismic datasets (\cite{seisbench}) thus often rely on measurements from specific spatial points, where those seismometers are installed, limiting their spatial coverage and resolution.
        Indeed, acquiring such dense, labeled datasets is infeasible with current physical seismometer deployments due to their sparse spatial coverage and high costs. More recent systems, such as \textit{Distributed Acoustic Sensing}, are explored by geologists, but these are still in the early stage of deployment and cannot yet be used for generating large datasets (\cite{das_data}). Without sufficient real-world data to accurately reconstruct seismic event distributions, ML techniques often underperform, as they lack the granularity required to model complex seismic behaviors effectively.
        The question of how many observations are necessary to tackle seismological questions with ML techniques remains open. More specifically, the notion of \textit{data budgeting} (\cite{zhao_data_2022}), namely, the \textit{final performance prediction} and the \textit{needed amount of data prediction} are two questions that need to be answered to deploy ML architectures in real-life settings, limiting the costs of the measurement devices deployment.
        We present \texttt{PyAWD}, a Python library designed specifically for generating synthetic seismic datasets with high spatial and temporal resolution, as a tool for easily generating simulations whose results are directly usable by ML frameworks, and we show its usefulness to study the question of \textit{epicenter retrieval} and three related \textit{data budgeting} questions. This tool allows researchers to produce data that would currently be challenging to obtain in the real world, enabling investigations on how sparse real-world measures can be to solve seismic inverse problems. \texttt{PyAWD} uses customizable simulations of the acoustic wave propagation (\cite{demanet2015topics}) through 2D and 3D heterogeneous fields, and creates PyTorch-compatible datasets that researchers can easily deploy in ML pipelines. The generated datasets implement the standard PyTorch Dataset interface, ensuring seamless integration with PyTorch DataLoaders. They also support on-the-fly data generation and augmentation, critical for training robust models. This format enables efficient data loading, batching, and shuffling during training. It also allows seamless use of GPU acceleration and parallel data processing through PyTorch DataLoaders (\cite{pytorch}).
        Being based on simulated data, \texttt{PyAWD} does, of course, not solve the problem of the lack of labelled real-life data. However, we claim that this tool can be used to determine how many observations would be necessary to achieve some given accuracy saturation (\cite{learning_curve}), to make reliable decisions.
        
        Recent advancements in Machine Learning (ML) have demonstrated potential for applications such as earthquake prediction and epicenter localization, improving inference accuracy and speed (\cite{review}). However, the limited availability and low spatial resolution of current seismic datasets constrain further progress. As noted by (\cite{hemew3d}), there is a critical demand for high-resolution spatio-temporal data to fully harness ML's predictive capabilities in seismology. \texttt{PyAWD} offers a solution by generating synthetic, high-resolution datasets that emulate waves' behaviors through diverse geological conditions, facilitating robust feature extraction and pattern recognition. This approach enables the creation of datasets easily pluggable in neural networks training pipelines, making \texttt{PyAWD} a foundational tool for developing advanced ML models for seismic analysis.
        Previous studies, such as (\cite{review2, hemew3d}), highlight the scarcity of synthetic datasets focused on seismic analysis for Machine Learning. For example, satellite imagery has been used for surface deformation studies related to earthquakes (\cite{quakeset}), and seismometer data has been gathered at both local (\cite{pnw}) and global scales (\cite{stead}). However, these datasets lack the spatial coverage and propagation field representations essential for ML. 
        There are existing tools, like \textit{Synthoseis}, that provide 2D simulations (\cite{synthoseis, deepwave}), but, to our knowledge, no solution offers large sets of comprehensive annotated 2D and 3D simulations with high spatio-temporal resolution. \texttt{PyAWD} fills this need, enabling research with enriched data coverage for ML applications.
        \texttt{PyAWD} offers several features that make it a valuable asset for geophysicists and data scientists:
        \begin{itemize}
            \item Customization: Users can define simulation parameters, such as material properties, simulation durations, spatial and temporal discretization, and source characteristics.
            \item Comprehensive spatio-temporal coverage: Data can be accessed through seismograms at selected spatial points or as a complete spatial overview, allowing the study of questions not yet easily accessible through real-world experiments.
            \item Integration with ML frameworks: By offering PyTorch-compatible datasets, \texttt{PyAWD} facilitates seamless ML pipeline integration, enabling the development of accurate, spatially informed models, annotated with learning curves showing the impact of the data budget on the attainable accuracy, for earthquake prediction and other geophysical applications.
        \end{itemize}
        The first two assets are direct consequences of the use of simulated data. Our library adds the convenience of use and the automation of the generation of various initial conditions, a needed characteristic when dealing with Monte Carlo simulations that existing tools do not offer. The last point of embedding the simulations directly into PyTorch datasets is also a novelty when compared to the state-of-the-art solutions.
        The paper is structured as follows: Section \ref{sec:materials} presents the methods and materials used in the development of \texttt{PyAWD}, including an in-depth discussion of the underlying algorithms and technical specifications. Section \ref{sec:results} reports the results of our simulations, detailing the performance metrics and comparing our tool with existing methods. Section \ref{sec:discussion} presents a toy example study, focused on the question of epicenter retrieval and its data budgeting, to show the usefulness of \texttt{PyAWD}, and placing our work in the context of related research, identifying limitations, and suggesting future directions.

\section{Methods/Materials and Methods}\label{sec:materials}
        \texttt{PyAWD} is a tool designed to facilitate the simulation of wave propagation phenomena in heterogeneous media and to assess the reliability of predictive ML models, by enabling the development  of ML pipelines and the study of data budgeting for many problems. It allows users to define and solve various forms of the acoustic wave equation with the flexibility to customize parameters such as wave speed, attenuation, propagation field structure, and external forces. One of the key features of \texttt{PyAWD} is its ability to handle spatio-temporally varying propagation fields, making it possible to simulate wave behavior in heterogeneous environments, such as geological formations with varying material properties. \texttt{PyAWD} is intended to be easily used with Deep Learning by producing PyTorch's Datasets, enabling the study of complex propagations through Machine Learning models.
        2D and 3D visualization capabilities are another strength of \texttt{PyAWD}, allowing users to generate and analyze representations of wave propagation behaviors. These visualizations include detailed 2D or 3D plots of the wave propagation in the field and at specific points, helping users interpret the results of their simulations. \texttt{PyAWD} includes tools for visualizing and interacting with \textit{interrogators}, which are specialized functions used to probe the simulated wave fields at specific points, simulating the behavior of traditional seismometers, but proposing an abstraction of the physical device.
        \texttt{PyAWD} uses the anisotropic nondispersive Acoustic Wave Equation (\cite{demanet2015topics}), which can be expressed as
            \begin{equation}\label{eq:awe}
                \frac{d^2u}{dt^2} = c\nabla^2 u - \alpha \frac{du}{dt}+ f
            \end{equation}
        in which $u$ is the displacement field (a vector field, typically measured in \textit{meters}, if the equation is used in its vectorial form, or a scalar field if the equation is used in its scalar form), $t$ is the time (scalar, in \textit{seconds}), $c$ describes the wave propagation speed field (a scalar field, in $\frac{s^2}{m^2}$), $\alpha$ is an attenuation factor (scalar, in $\frac{1}{s}$) and $f$ is an external force (a vector field, in $\frac{m}{s^2}, $if the equation is used in its vectorial form, or a scalar field otherwise). \texttt{PyAWD} implements both the scalar and the vectorial form.
        The vectorial 2D acoustic wave equation is discretized with the finite differences scheme in equation \ref{eq:awe2}. The derivation is similar for the 3D version.
        \begin{equation}
            \label{eq:awe2}
            \begin{aligned}
                &\left[
                \begin{matrix}
                \frac{\partial^{2}}{\partial t^{2}} u_x\left(t, x + \frac{h_x}{2}, y\right)\\
                \frac{\partial^{2}}{\partial t^{2}} u_y\left(t, x, y + \frac{h_y}{2}\right)
                \end{matrix}
                \right] = \\
                &\left[
                \begin{matrix}
                \left(\frac{\partial^{2}}{\partial x^{2}} u_x\left(t, x + \frac{h_x}{2}, y\right) 
                + \frac{\partial^{2}}{\partial y^{2}} u_x\left(t, x + \frac{h_x}{2}, y\right)\right) c(x, y, t) 
                + f_x\left(t, x + \frac{h_x}{2}, y\right)\\
                \left(\frac{\partial^{2}}{\partial x^{2}} u_y\left(t, x, y + \frac{h_y}{2}\right) 
                + \frac{\partial^{2}}{\partial y^{2}} u_y\left(t, x, y + \frac{h_y}{2}\right)\right) c(x, y, t) 
                + f_y\left(t, x, y + \frac{h_y}{2}\right)
                \end{matrix}
                \right]
            \end{aligned}
        \end{equation}
        The wave propagation simulations are run using Devito (\cite{devito-compiler, devito-api}), a finite difference partial differential equations (PDE) solver library written in Python. Devito is an open-source framework designed for solving PDE using the finite difference method. It allows users to express finite difference problems symbolically using Python, specifically the SymPy library (\cite{sympy}), enabling them to define complex mathematical operators and equations in concise symbolic notation. This high-level symbolic representation allows users to focus on algorithmic development rather than low-level implementation details. Devito in itself already requires some expertise and deep mathematical understanding. Therefore, the abstraction of the practical numerical solving is highly desirable for seismologists and ML engineers. \texttt{PyAWD} offers this abstraction. However, other equations can easily be implemented and added to the library. The results of the simulations are embedded in PyTorch datasets. Since the use of neural networks for signal analysis is common in seismic signals analysis, such as earthquake detection (\cite{detection}), phase picking (\cite{phase-picking-cnn, phase-picking-attention}), signal denoising (\cite{denoising}), location (\cite{location}), and structure identification and classification (\cite{structure-classification}), and most neural network applications require heavy learning procedures, for which optimized data handling is needed, Deep Learning frameworks stand as principal actors in current Machine Learning. PyTorch (\cite{pytorch}) is one of the most popular neural network libraries. It allows the design of highly scalable models, optimized for GPU and clusters. This library benefits from an active community. Providing a dataset generator in PyTorch allows using diverse efficient tools such as \texttt{Pytorch DataLoaders} and \textit{torchvision transformations} (\cite{torchvision2016}) for the processing of the data. As PyTorch tensors are seamlessly castable to NumPy arrays, all the available functions of this environment (including \texttt{SciPy} and \texttt{sklearn}) are accessible. PyTorch offers a comfortable bridge between basic Python code and highly scalable models.
        
        The form of equation \ref{eq:awe} allows the generation of many simulation scenarios. As an example, the parameter $c$ in equation \ref{eq:awe}, representing the wave propagation speed, introduces the concept of spatio-temporal heterogeneity into the simulation. This parameter can vary with spatial and temporal coordinates, allowing us to define it as a function $c(x, y, t)$. \texttt{PyAWD} is designed to enable users to define their propagation fields, providing significant flexibility in modelling complex systems. By allowing $c$ to be a function of both space and time, we can simulate environments with varying material properties, such as geological formations with different rock densities or temperatures. This capability is crucial for accurately capturing the physical phenomena of wave propagation in heterogeneous media, which is often encountered in real-world applications. 
        
\begin{figure}
            \centering
            \begin{subfigure}[b]{0.75\textwidth}
                \centering
                \includegraphics[width=\linewidth]{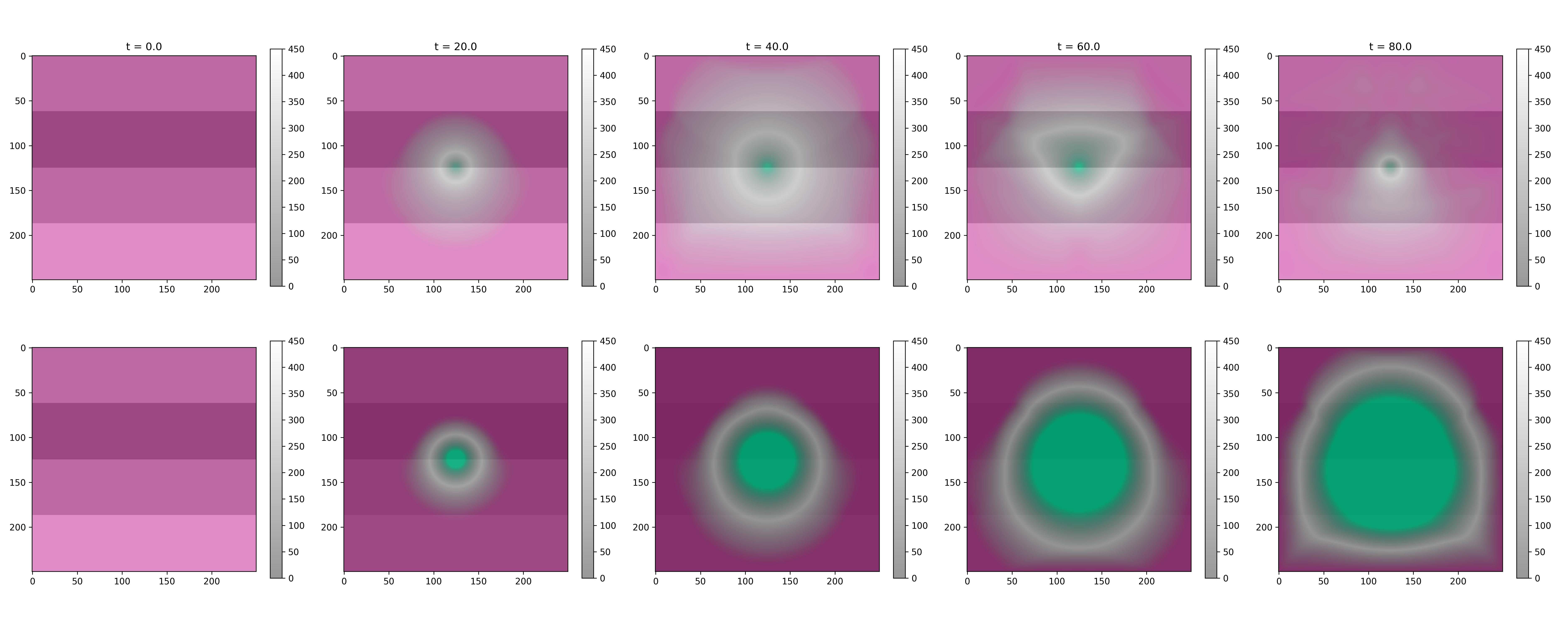}
                \caption{Scalar wave propagation. The first line shows the propagation of the wave in a spatially varying propagation speed field that remains temporally constant. The second row shows a propagation in the same initial field, but where the field is scaled by a temporally varying factor $A$, such that $c(x, y, t) = A(t)\times c(x, y, 0)$.}
                \label{fig:varyingc1}
            \end{subfigure}
            \hfill
            \begin{subfigure}[b]{0.2\textwidth}
                \centering
                \includegraphics[width=\linewidth]{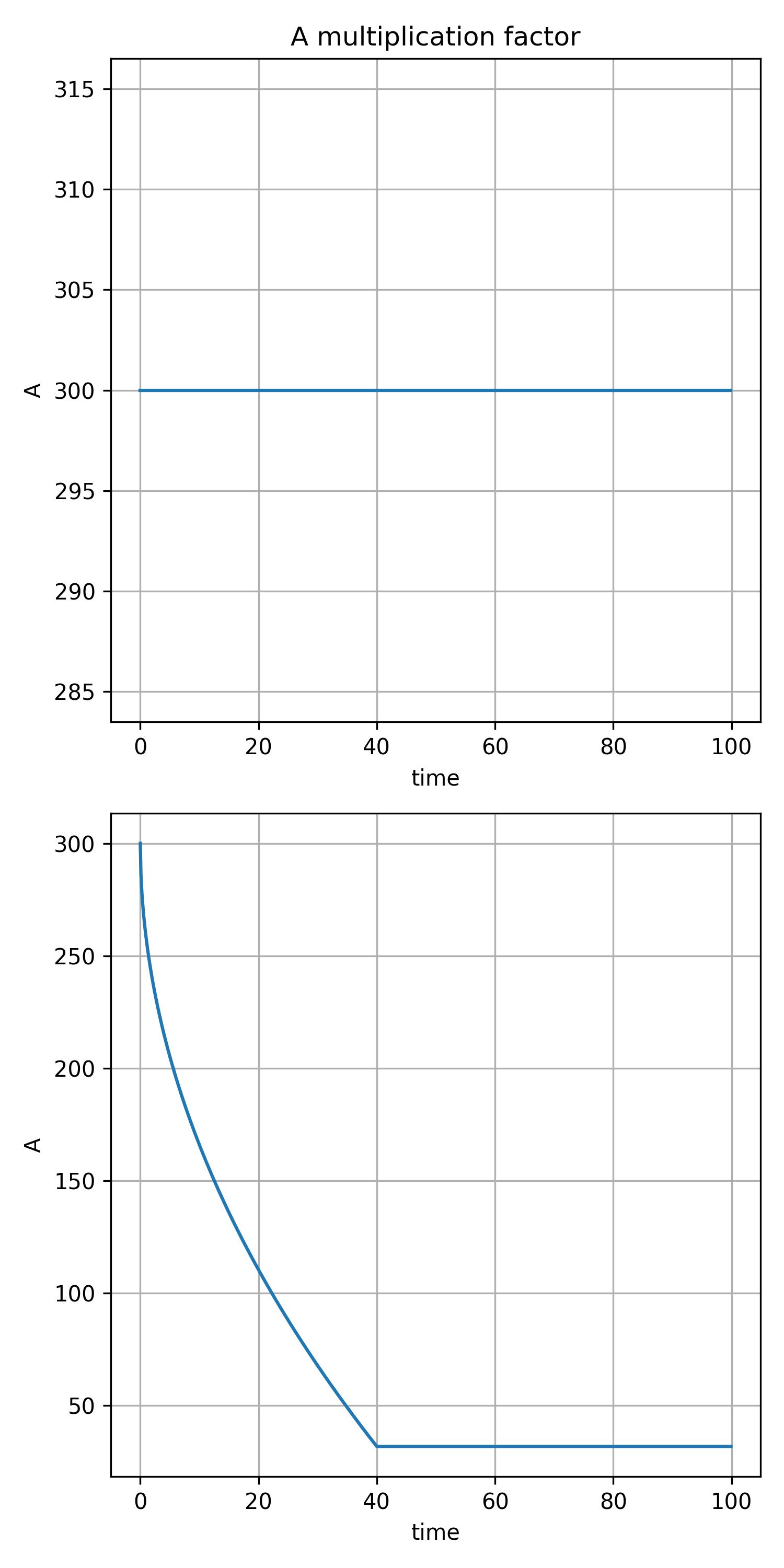}
                \caption{Variation of the factor $A$ for both the constant and the decreasing case.}
                \label{fig:varyingc2}
            \end{subfigure}
            \caption{Example of scalar wave propagation simulation generated by \texttt{PyAWD}.}
            \label{fig:varyingc}
        \end{figure}

\begin{figure}
            \centering
            \includegraphics[width=0.4\textwidth]{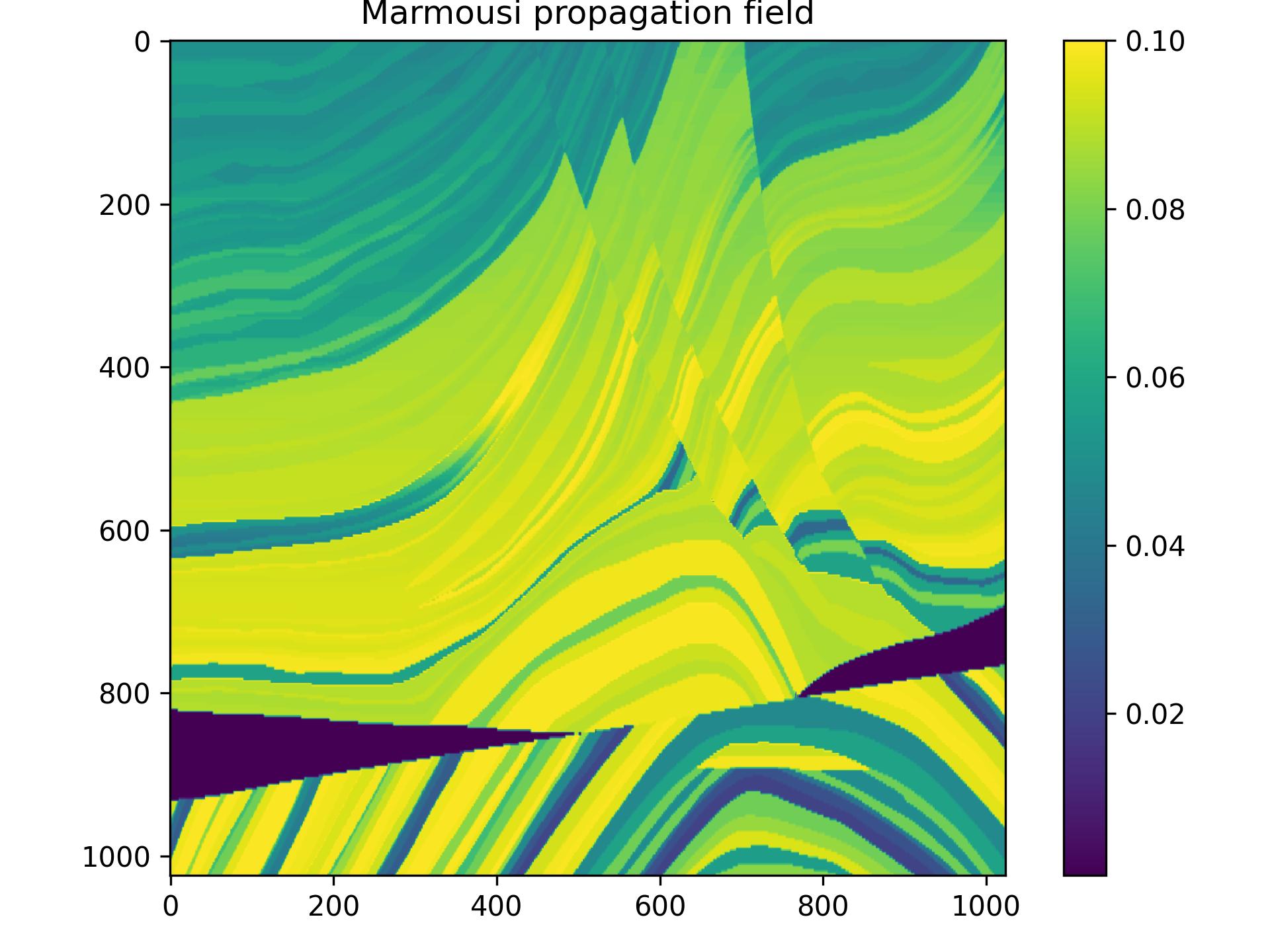}
            \caption{The Marmousi field, from (\cite{marmousi}). The darker the color, the lower the propagation speed. This complex structure is used as a preset example in \texttt{PyAWD}.}
            \label{fig:marmousi}
        \end{figure}

        An example of such behavior is illustrated in  \ref{fig:varyingc}, in which the propagation speed progressively decreases (which is a way of modeling a temperature increase). The equivalent propagation in a temporally constant propagation field is given in the same figure. In both figures, the initial wave propagation velocity field is the same, and is made of four horizontal bands illustrating different materials through which the wave will travel. In the first line of simulation, the field stays constant through the simulation. In the second simulation, the initial field is scaled by a factor shown on the last plot. In those figures, darker bands mean a smaller speed of propagation. Those are expressed in transparency. The white, rose, and green colors denote the scalar value of the displacement field.
        Then, an example of a propagation field is given in figure \ref{fig:marmousi}. This example, the Marmousi field (\cite{marmousi}), is provided as a basic preset in \texttt{PyAWD} and shows an example of a highly discontinuous field that can be used to generate simulations. The darker the color, the smaller the propagation speed. In \texttt{PyAWD}, by default, this field is scaled by a random factor for each simulation. However, as we will do in section \ref{sec:results}, this factor can be kept constant.
        This flexibility allows for the exploration of a wide range of problems: retrieving the underlying structure of a medium from the seismic measurements, forecasting the propagation from early measurements,  equation retrieval, where the governing equations of a system are inferred from observed data are example of question that can be tackled using \texttt{PyAWD}.
        \label{sec:solver}
        As \texttt{PyAWD} targets interdisciplinary researchers, such as data scientists, geologists, seismologists, and volcanologists, we provide different \textit{Jupyter Notebooks}, along with the library itself, which introduce the different notions needed for the best utilization of the tool. Those include:
        \begin{itemize}
            \item An introduction to the Scalar and Vector Acoustic Wave Equation solution using Devito,
            \item An introduction to the use of \textit{interrogators},
            \item An introduction to the dataset generation,
            \item An example of heterogeneous fields usage,
            \item An example of a spatio-temporal varying field simulation
        \end{itemize}
        Those are meant to be comprehensive guides to the proper usage of \texttt{PyAWD}. Alongside those, we propose a complete documentation\footnote{\url{https://pascaltribel.github.io/pyawd/}} of the different classes and functions.

\section{Validation and results}\label{sec:results}
        In this section, we present a toy example of a use case for \texttt{PyAWD}. We focus on the concept of 2D-\textit{epicenter retrieval}, well-defined in (\cite{cnn_location}), which aims at determining the absolute coordinates $(x_0, y_0)$ of a wave epicenter, based on temporal measurements of ground motion at specific locations, referred to as \textit{seismograms}, where seismometers are placed. Traditionally, this problem is addressed using three or more seismometers (\cite{braile2002as1}). The measurements obtained from these devices allow for an estimation of the \textit{epicentral distance}, which can subsequently be used to triangulate the epicentral coordinates. However, when fewer than three measurement stations are available, the prediction is subject to uncertainty. Recent studies have demonstrated that Machine Learning techniques can extract valuable information from the available seismograms (\cite{cnn_location, improvement_baz, eqconvmixer, dl_epicenter}), yielding promising results. Indeed, we consider the hypothesis which states that the interference that the propagating waves causes with itself when reflecting on the propagation field heterogeneities can bring valuable information about the structure of this field, information that is, at least partially, contained in the measured signal. This hypothesis is supported by the way the human brain can accurately capture a sound source position even though the number of points at which the wave is measured is lower than three (\cite{audio_localisation}). We claim that this information can be retrieved and used to reduce the uncertainty around the computed epicenter. We note, secondly, that the task presents another difficulty when compared to a \textit{real-life} situation: \textit{elastic waves}, which are often considered to accurately model earthquakes' propagation, can be decomposed into two orthogonal waves (primary and secondary), which have different propagation speeds. The difference in arrival times for those two waves allows computing the distance between the earthquake epicenter and the seismometer position. However, in our use case, the wave is \textit{acoustic}: such distance information cannot be inferred directly from the signal, or at least, only the interference between the signal and itself could gather this information. We hypothesize that statistical/Machine Learning tools can extract this information and use it to retrieve the epicenter coordinates.
        \subsection{Data generation}
\begin{figure}
            \centering
            \includegraphics[width=\textwidth]{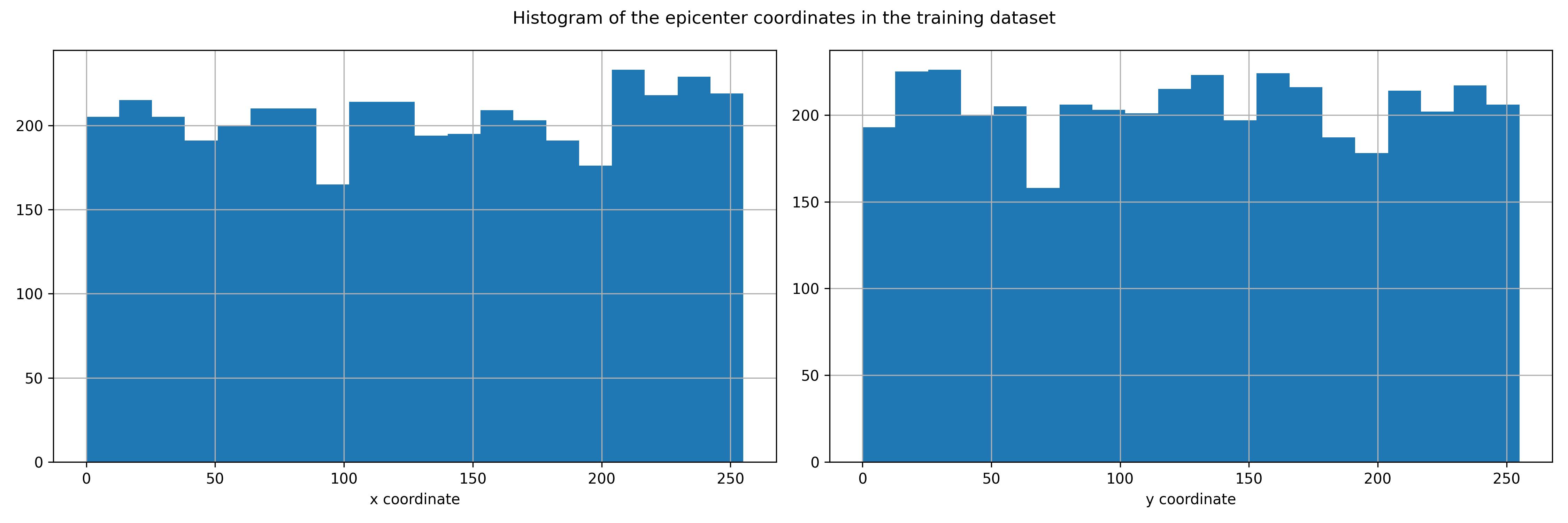}
            \caption{Histogram of both $x$ and $y$ coordinates of the epicenters in the training set.}
            \label{fig:histogram_epicenters}
        \end{figure}

\begin{figure}
            \centering
            \includegraphics[width=\textwidth]{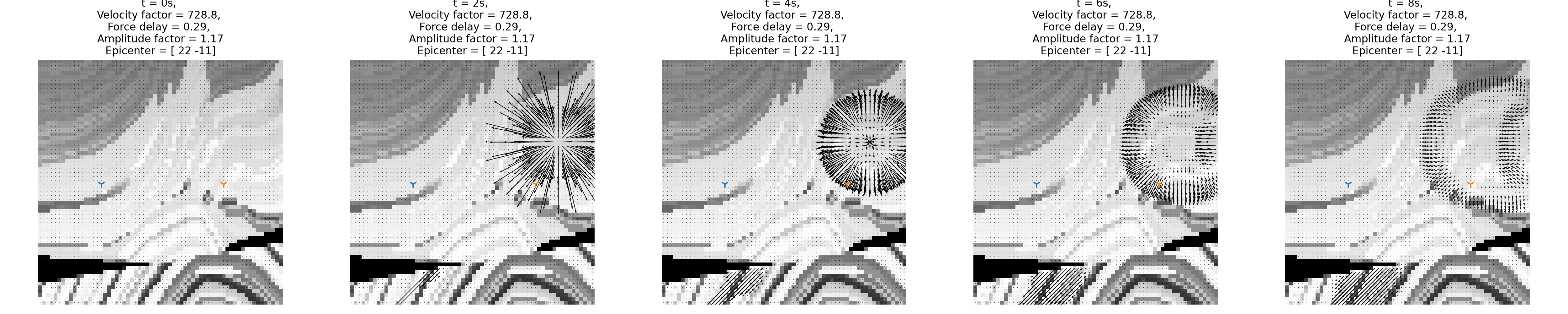}
            \caption{Example of propagating wave in the Marmousi field, with two interrogators.}
            \label{fig:train_example}
        \end{figure}

\begin{figure}
            \centering
            \includegraphics[width=0.9\textwidth]{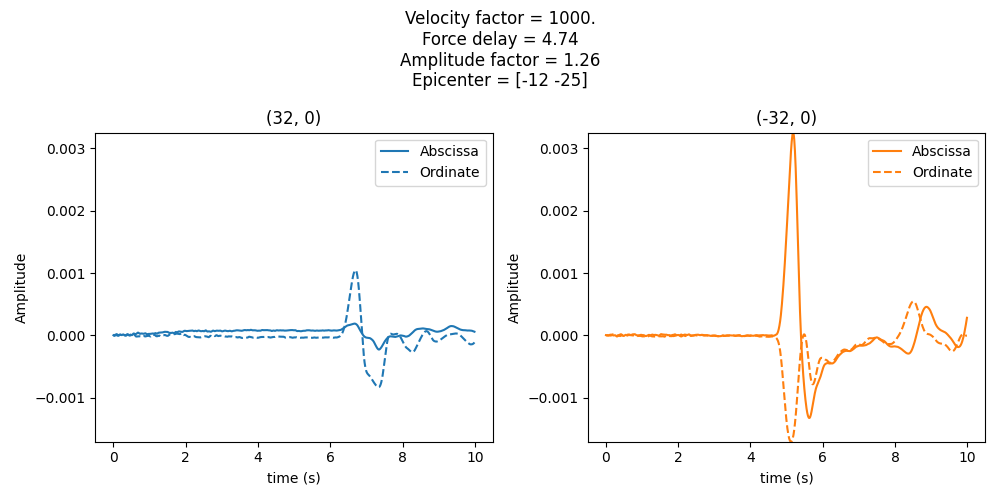}
            \caption{Example of interrogators response, which follows the simulation shown in figure \ref{fig:train_example}}
            \label{fig:train_interrogators_example}
        \end{figure}

        The experiment aims at assessing that \texttt{PyAWD} does not oversimplify reality, by showing a concrete problem whose solution is non-trivial. We employ \texttt{PyAWD} to generate a dataset consisting of 2D acoustic wave propagation simulations in the Marmousi field (see figure \ref{fig:marmousi}), a well-known and widely used example of a heterogeneous propagation field. Two datasets are created, one for training and another for testing, containing $3584$ and $512$ simulations, respectively. The field is discretized into a $256 \times 256$ grid. Each simulation runs for $10$ seconds, with two interrogators placed at coordinates $(-64, 0)$ and $(64, 0)$ recording the ground motion in both the vertical and horizontal planes at a frequency of $100 \hertz$. The epicenter coordinates, the external force amplitude, and the external force delay are randomly selected, respectively, from within the propagation field, the interval $[0.5, 1.5[$, and the interval $[0, 5[$ seconds. The size of the total dataset is chosen such that the proportion of epicenters present in the training dataset (number of epicenter coordinates used divided by the number of possible coordinates in the discretized space) remains small: $\frac{3584}{256\times 256} = 5.57\%$. This proportion has to be considered along with the fact that the aforementioned parameters are random as well and add diversity to the data. 
        The complete (training and testing together) dataset is therefore made of $4096$ elements, each one made of $2\times2$ features (horizontal and vertical movement for each interrogator), and each feature is a $1000$-timesteps long vector. In the following study, no feature extraction is performed. We follow the previous hypothesis that the information required to compute the regression of the epicenter coordinates is contained directly in the measured signal, which results from the interactions the wave has with itself when crossing different heterogeneities on its path, and that this information would be lost when applying statistical feature extraction. The raw seismograms are used \textit{as are} directly in the different architectures. The epicenter coordinates are sampled out of a uniform distribution and the histograms of the epicenter coordinates in the training set are shown in figure \ref{fig:histogram_epicenters}. An example of item from the dataset is shown in figure \ref{fig:train_interrogators_example}, where the measurements correspond to the simulation shown in figure \ref{fig:train_example}.
        \subsection{Machine Learning pipeline}
            The dataset is randomly split into three subsets. First, we split the dataset into two parts: a \textit{training} set (3284 samples) and a \textit{testing} set (512 samples). Then, during the training procedure, which is repeated 10 times, we randomly split the \textit{training} set into a subset used for model optimization and a \textit{validation} set (respectively 80\% and 20\% of the initial training set). Each model is then optimized on the training set in a supervised manner, using the validation set as a generalization performance assessment tool. Finally, for each of the 10 experiments, the trained models generate their predictions for the \textit{testing} dataset. 
            
            We attack the regression problem using different Machine Learning strategies. We use 7 different Machine Learning architectures, taking as input the raw seismograms measured by the interrogators. Those include different families of architectures, in increasing order of complexity: 
            \begin{itemize}
                \item Baseline: Constant model (here, the constant is the average of the coordinate vectors in the training dataset)
                \item Ridge Regression
                \item KNN Neighbors
                \item Decision Tree (\cite{breiman1984classification})
                \item Extra Trees (\cite{geurts2006extremely})
                \item Neural Networks:
                    \begin{itemize}
                        \item Multi-layer Perceptron (\cite{hinton1989connectionist})
                        \item Temporal Convolutional Neural Network (\texttt{TCNN}) (\cite{lea2016temporalconvolutionalnetworksaction})
                    \end{itemize}
            \end{itemize}
            All of them are implemented in \texttt{sklearn} (\cite{sklearn}) except for the \texttt{TCNN}, which is implemented in PyTorch (\cite{pytorch}).

            Since the epicenter coordinates are vectors of two components, the regression task is a multitask problem. All the presented methods are therefore duplicated to compute separately the two coordinates, except for the TCNN whose output is directly a two-component vector.
            
            The models are evaluated on the \textit{testing} dataset using the \textit{normalized mean squared error} (\texttt{NMSE}) between the expected and predicted epicenter coordinates vector:
            \begin{equation*}
                \text{NMSE} = \frac{\sum_{i=1}^{N}(y_i - \hat{y_i})^2}{\sum_{i=1}^{N}(y_i - \bar{y})^2}
            \end{equation*}
            where
            \begin{equation*}
                \bar{y} = \frac{\sum_{i=1}^{N}y_i}{N}
            \end{equation*}
            is the average coordinate vector in the training dataset, and $N$ is the number of samples in the testing set.
        
        \subsection{Epicenter retrieval: results}
\begin{figure}
                \centering
                \includegraphics[width=0.8\linewidth]{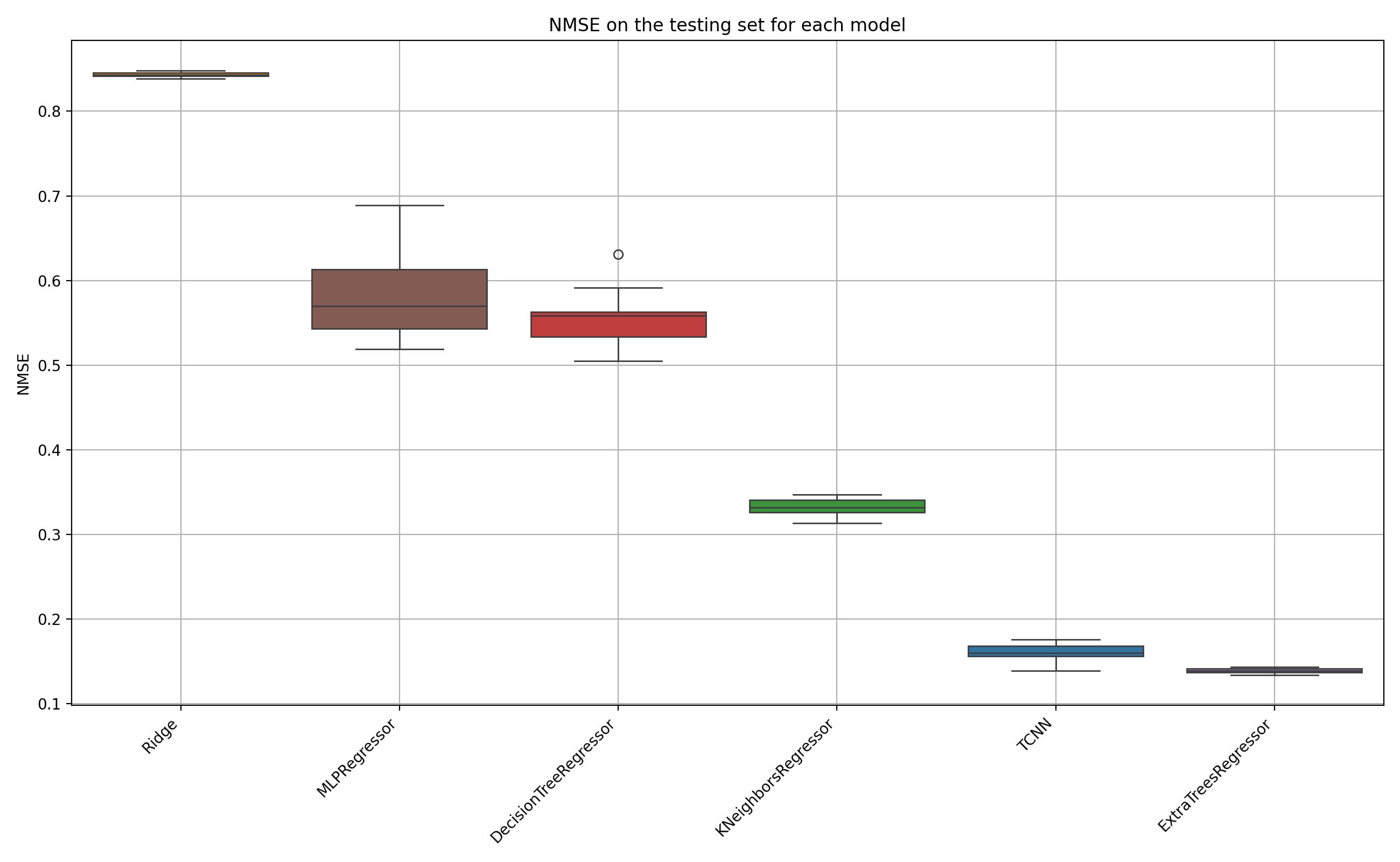}
                \caption{Normalized Mean Squared Error of the predicted coordinates for each model. The training/testing procedure has been done 10 times on the randomly split dataset.}
                \label{fig:NMSES}
            \end{figure}

            Figure \ref{fig:NMSES} shows the normalized mean squared error of the predictions of each model on the testing set, ordered by decreasing value.
            
\begin{figure}
                \centering
                \begin{subfigure}[b]{0.45\textwidth}
                    \centering
                    \includegraphics[width=\linewidth]{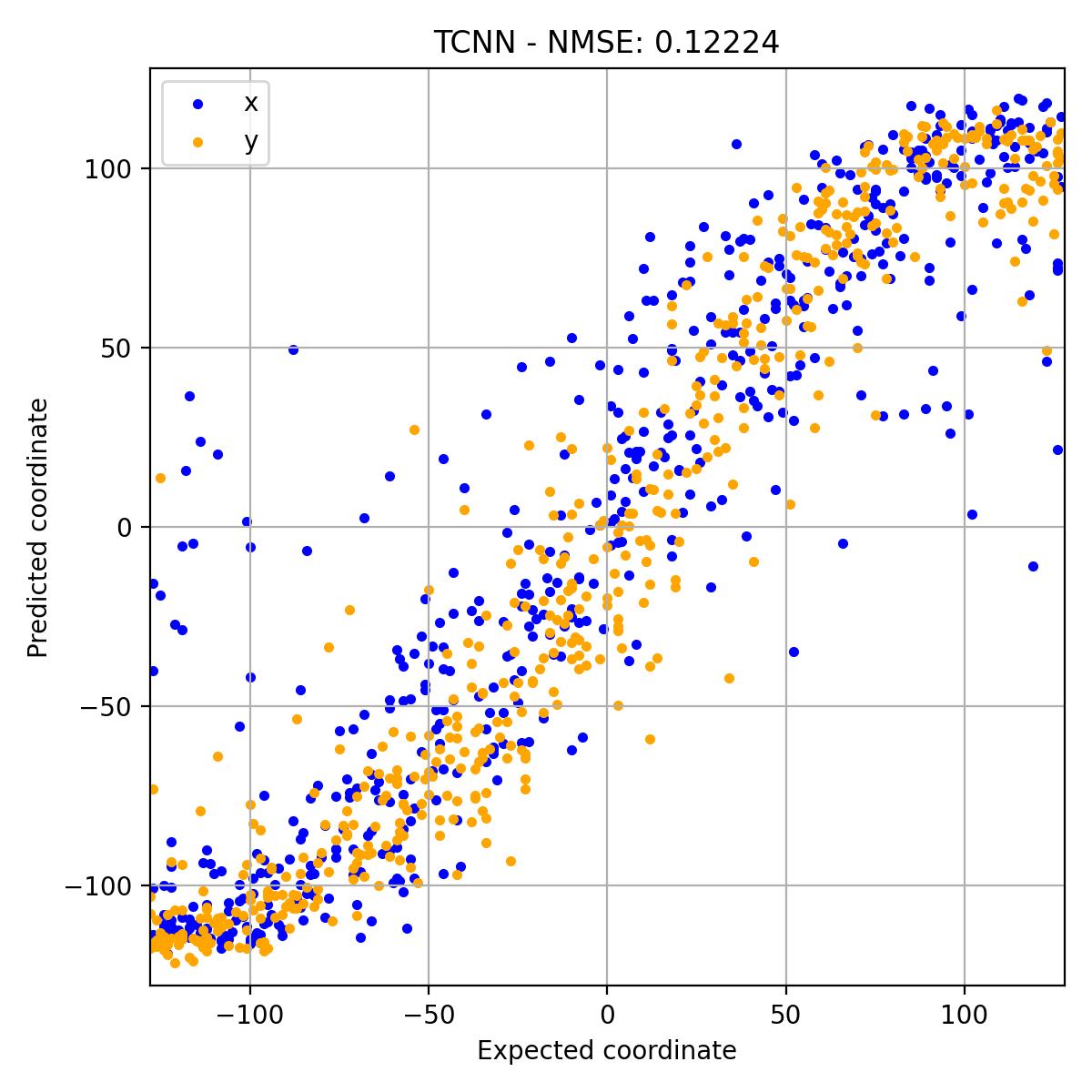}
                    \caption{Predicted versus expected coordinates for the testing set, using the Temporal Convolutional Neural Network.}
                    \label{fig:results_TCNN}
                \end{subfigure}
                \hfill
                \begin{subfigure}[b]{0.45\textwidth}
                    \centering
                    \includegraphics[width=\linewidth]{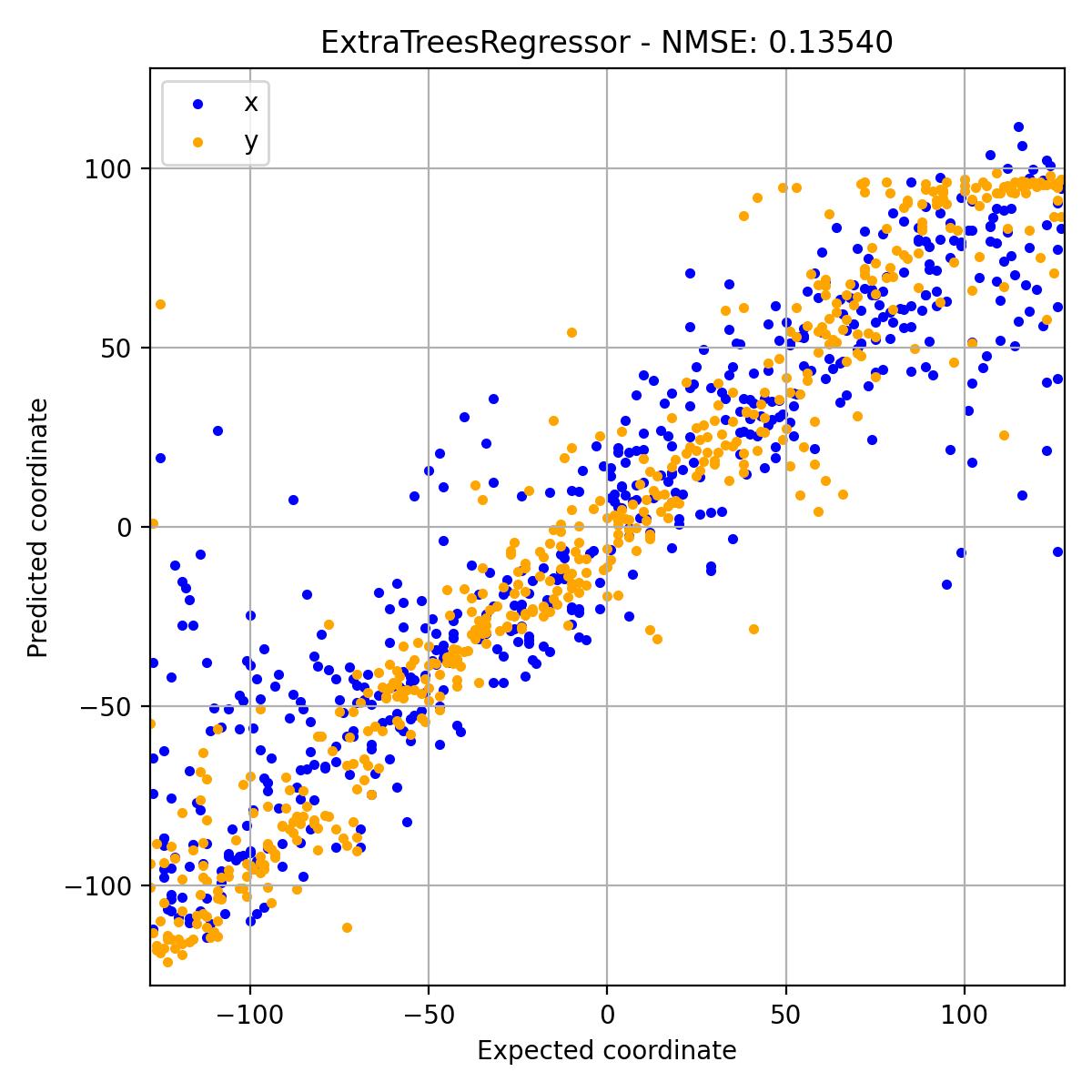}
                    \caption{Predicted versus expected coordinates for the testing set, using the ExtraTrees.}
                    \label{fig:results_ExtraTrees}
                \end{subfigure}
                \caption{Comparison of predicted versus expected coordinates for the testing set, using the Temporal Convolutional Neural Network and the ExtraTrees models. The blue points correspond to the horizontal coordinate, while the orange points correspond to the vertical coordinate.}
                \label{fig:results_comparison}
            \end{figure}

            First, figures \ref{fig:results_TCNN} and \ref{fig:results_ExtraTrees} show the predicted versus expected coordinates for both the TCNN and Extra Trees. The blue points correspond to the horizontal coordinate, while the orange points correspond to the vertical coordinate.

\begin{figure}
                \centering
                \begin{subfigure}[b]{0.45\textwidth}
                    \centering
                    \includegraphics[width=\linewidth]{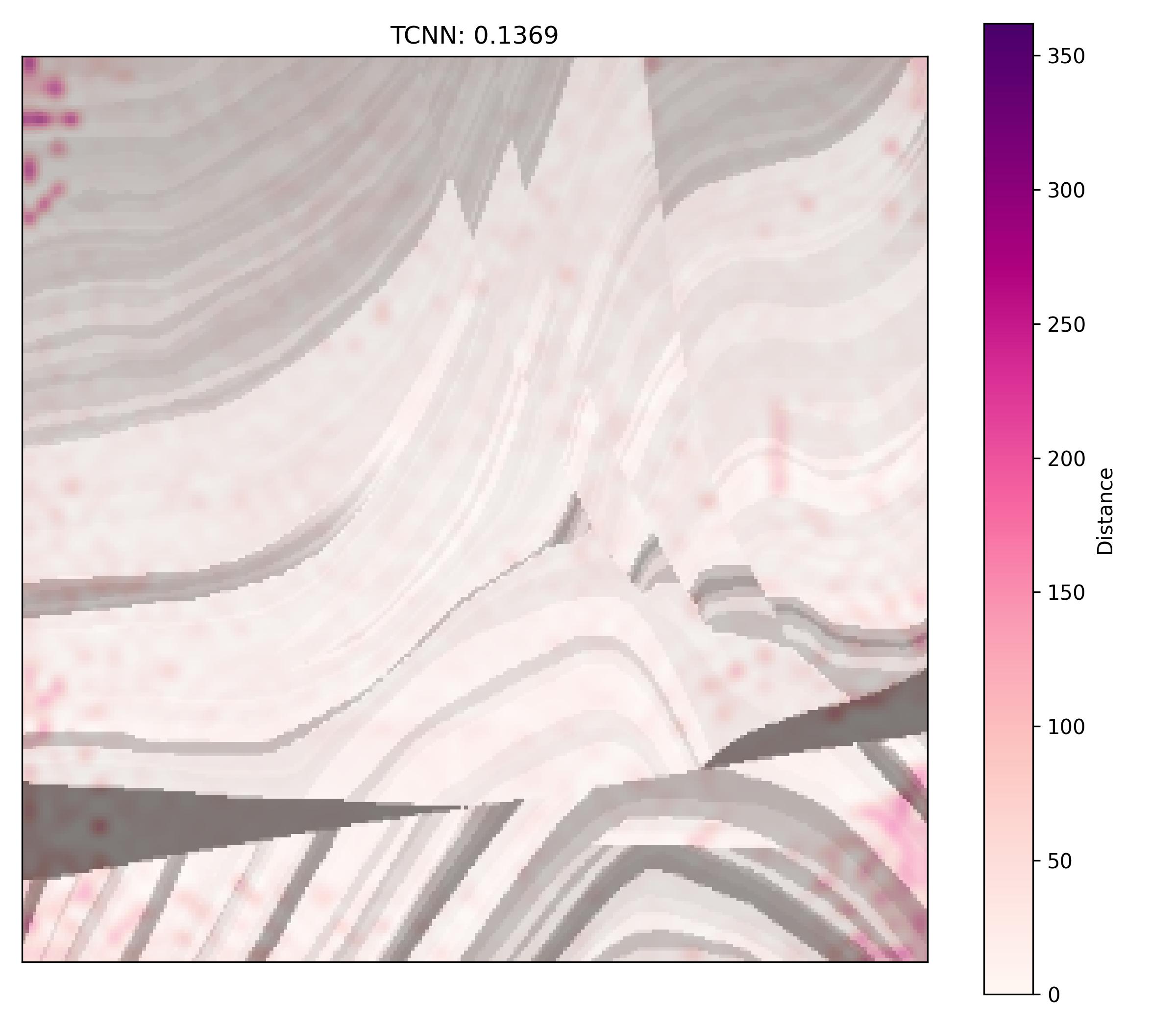}
                    \caption{Temporal Convolutional Neural Network}
                    \label{fig:heatmap_TCNN}
                \end{subfigure}
                \hfill
                \begin{subfigure}[b]{0.45\textwidth}
                    \centering
                    \includegraphics[width=\linewidth]{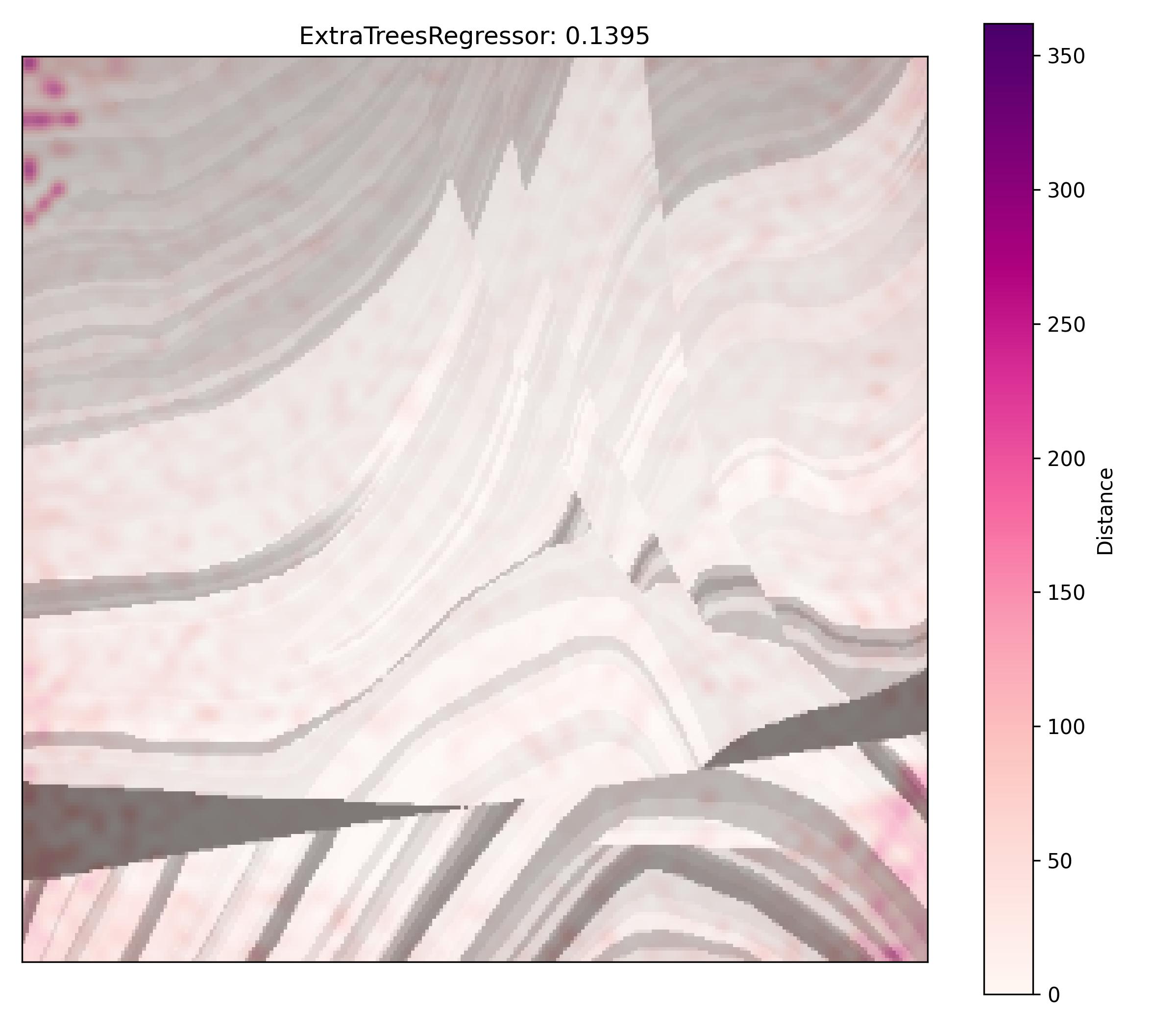}
                    \caption{ExtraTrees.}
                    \label{fig:heatmap_ExtraTrees}
                \end{subfigure}
                \caption{Heatmap of the distance between the expected and predicted coordinates for every possible epicenter. A darker color means that a wave starting in this location is less accurately predicted. The predictions are averaged on 10 different instances of each model.}
                \label{fig:heatmap_comparison}
            \end{figure}

            \texttt{PyAWD} can then be used to generate simulations and measurements from every possible coordinate in the propagation field. By measuring the distance between the actual epicenter and the predicted one for every coordinate, we can discover in which parts of the fields the different models show the greatest weakness. Figure \ref{fig:heatmap_comparison} shows, both for the Temporal Convolutional Neural Network and Extra Trees, the spatial distribution of the errors between the expected and predicted epicenters. For the sake of computational time, the grid on which the epicenters are drawn is downsampled by a factor of $8$ (meaning that every discrete position out of $8$ is used as an epicenter location). The lighter the color, the smaller the distance between the expected and predicted epicenter. The errors heatmap is superposed to the propagation field (plot in black and white), which is the Marmousi field presented in figure \ref{fig:marmousi}. This allows understanding the regions in which the errors are the greatest.

\begin{figure}
                \centering
                \begin{subfigure}[b]{0.45\textwidth}
                    \centering
                    \includegraphics[width=\linewidth]{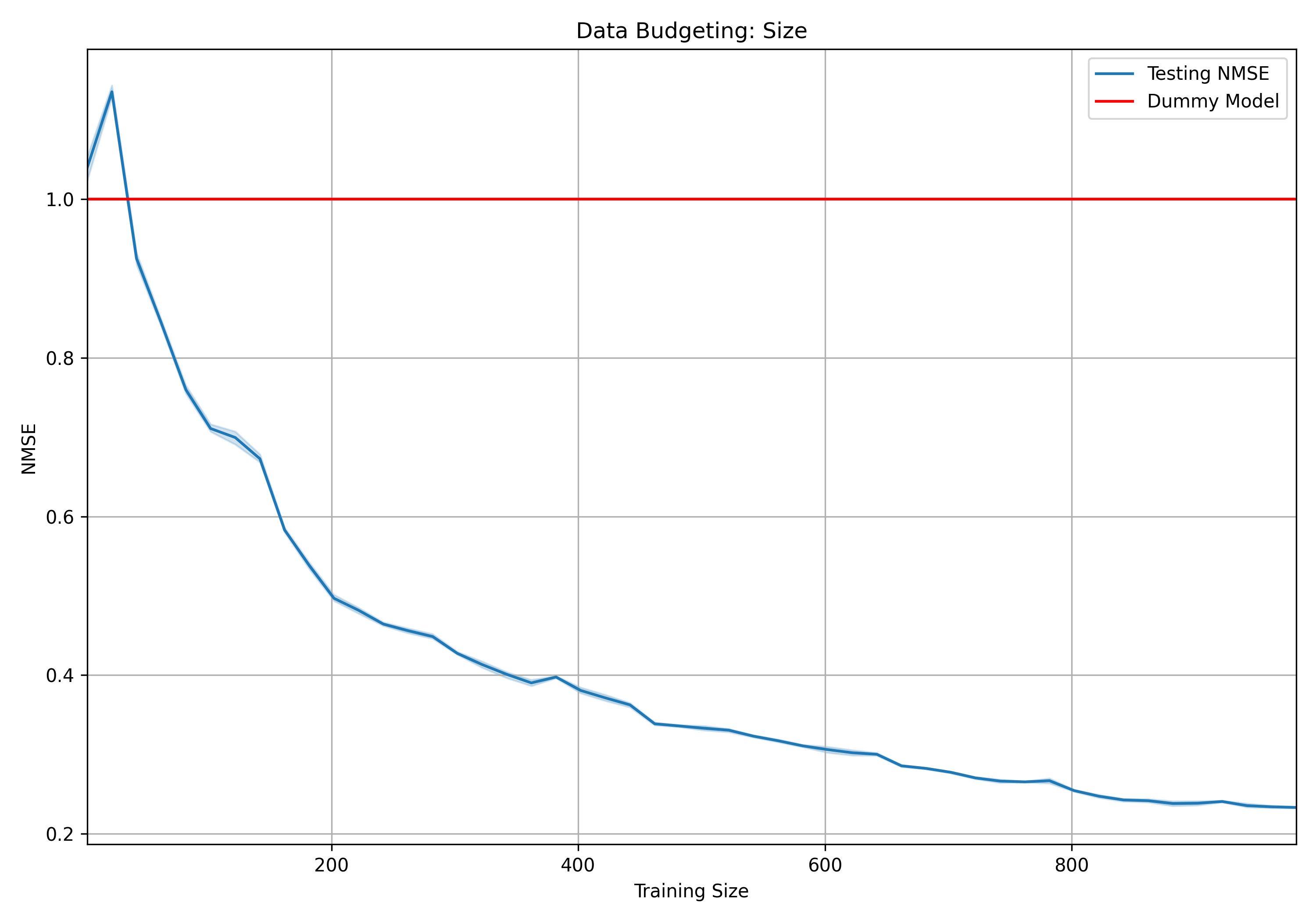}
                    \caption{NMSE depending on the number of samples in the training dataset. The red line shows the NMSE of a model predicting the average of the training set outputs.}
                    \label{fig:data_budgeting_size}
                \end{subfigure}
                \hfill
                \begin{subfigure}[b]{0.45\textwidth}
                    \centering
                    \includegraphics[width=\linewidth]{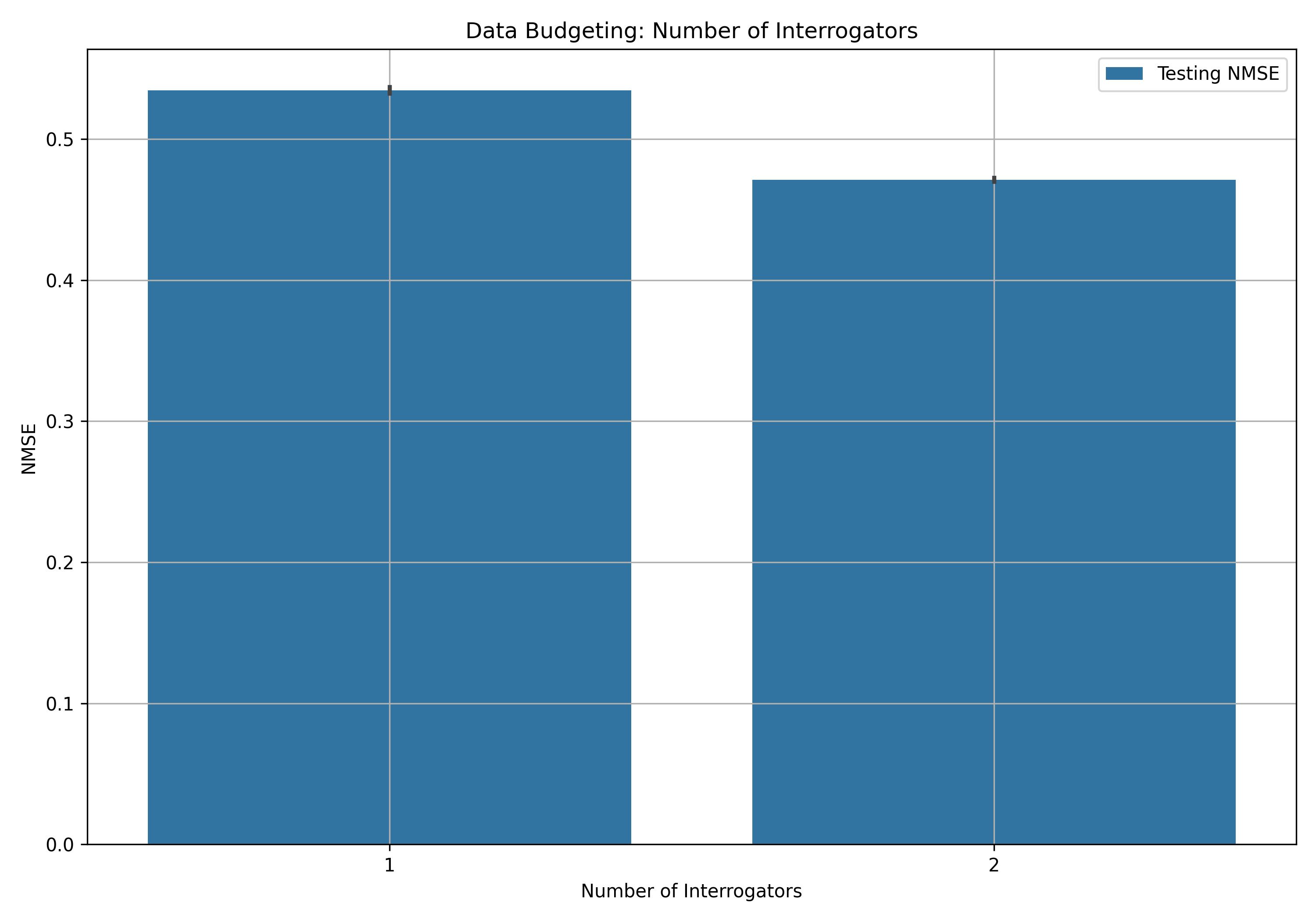}
                    \caption{NMSE depending on the number of interrogators used to generate the samples.}
                    \label{fig:data_budgeting_interrogators}
                \end{subfigure}
                \begin{subfigure}[b]{0.45\textwidth}
                    \centering
                    \includegraphics[width=\linewidth]{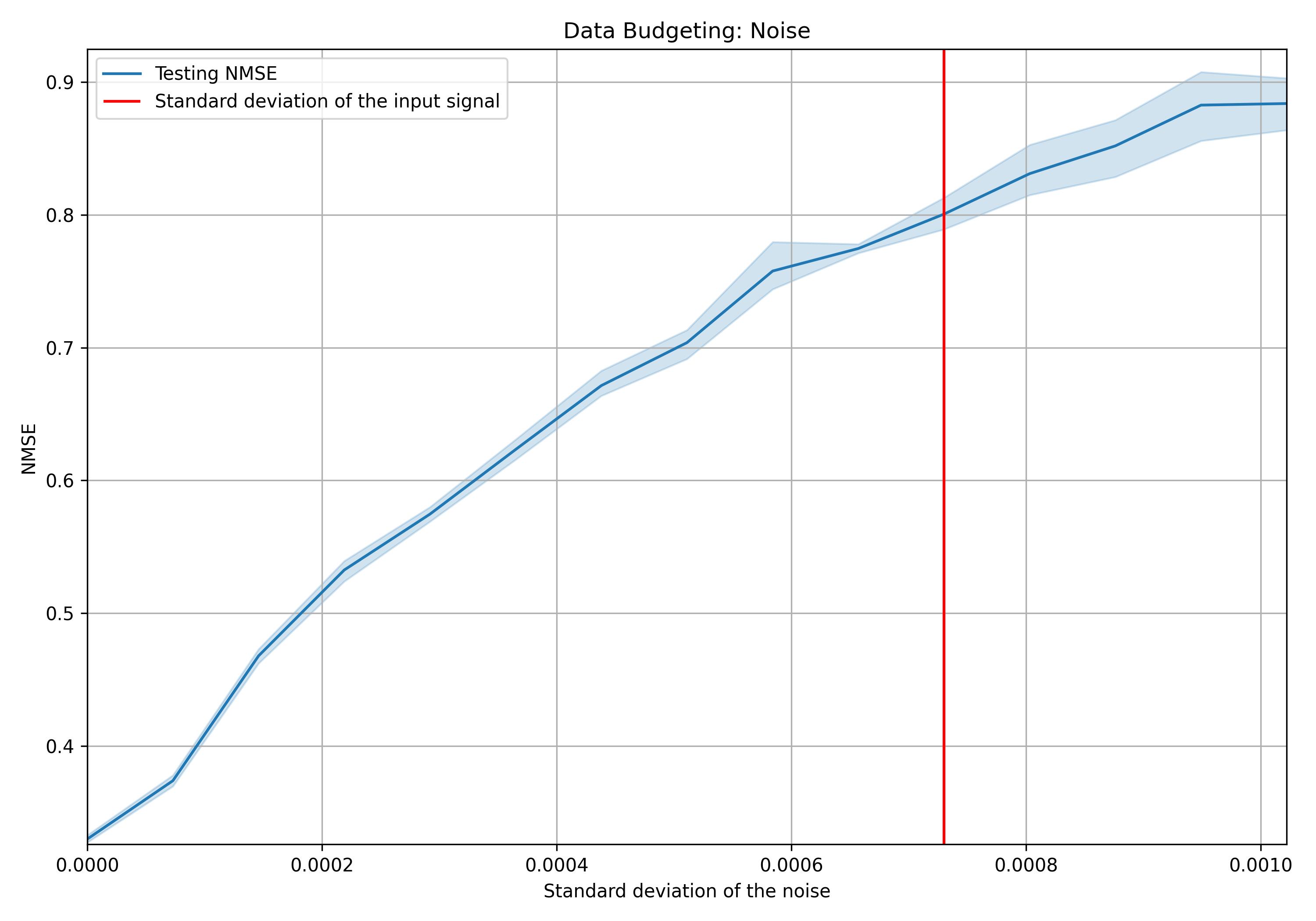}
                    \caption{Sensitivity to the noise. The red line shows the standard deviation of the input samples in the clean dataset, as a measure of the spread of the input signals.}
                    \label{fig:data_budgeting_noise}
                \end{subfigure}
                \caption{Data budgeting and noise sensitivity study. The predictions are averaged over 5 different instances of ExtraTrees. The confidence interval is $95\%$.}
                \label{fig:data_budgeting}
            \end{figure}

            Finally, the datasets generated by \texttt{PyAWD} can be used to tackle the problem of \textit{data budgeting}. We tackle this problem under three forms: first, we study the evolution of the NMSE depending on the number of samples in the training dataset. This evolution is shown in figure \ref{fig:data_budgeting_size}. Then, we study the NMSE depending on the number of interrogators used for generating the samples in the dataset. The results are presented in \ref{fig:data_budgeting_interrogators}. We finally show the sensitivity of the architecture to noise in the input data. We show how the NMSE evolves depending on the standard deviation added to the input data in figure \ref{fig:data_budgeting_noise}. We consider the addition of noise on the input data only to represent a real-life case where the measuring devices can be inaccurate and noisy.
    \section{Discussion}\label{sec:discussion}
        We first discuss the results of the practical example study of epicenter retrieval. We see that the Temporal Convolutional Neural Network and the Extra Trees largely outperform the other models. We therefore analyze their performances more in depth. However, a similar analysis can be done on the other ones. The spatial distribution of errors shown in figure \ref{fig:heatmap_comparison} shows the global quality of the model predictions, while showing larger errors closer to boundaries.   
    
        The data budgeting study highlights interesting results. First, $62$ samples in the training dataset seem to be enough to beat the constant average model. Moreover, the difference in NMSE between single-station and two-station predictions is marginal. Finally, we see that the model still achieves a NMSE lower than $1$, even if the standard deviation of the noise added to the input data is slightly larger than the standard deviation of the input data itself.
    
        The presented results show a practical use of \texttt{PyAWD} for studying the problem of epicenter retrieval, alongside the data budgeting aspect of the problem. 

        We then discuss the general aspects of \texttt{PyAWD} presented in this article.
        The results presented in this paper demonstrate the potential of \texttt{PyAWD} as a simulation tool that facilitates the use of Machine Learning approaches for wave analysis. By simulating wave propagation and providing a framework for integrating Machine Learning models, we have illustrated the usefulness of the tool specifically in the context of epicenter retrieval, with fewer than three interrogators. Through experiments comparing different models, we found that more complex architectures like Temporal Convolutional Neural Networks (TCNNs) and Extra Trees excel in handling spatio-temporal data, showcasing the need for advanced ML models in wave analysis.
        
        Overall, \texttt{PyAWD} serves as a robust and flexible solution for generating synthetic wave propagation datasets, which may reduce the cost and logistical demands associated with deploying numerous physical measuring devices in the real world to gather data.
        Our analysis highlights that \texttt{PyAWD} bridges a significant methodological gap in seismic wave analysis by facilitating the generation of high-quality synthetic datasets tailored to various geological configurations and wave characteristics. Traditional seismometer-based approaches often face logistical and data sparsity challenges, which can limit the robustness of analyses. \texttt{PyAWD} mitigates this issue by enabling fully customizable, high-resolution simulations that improve the diversity and quality of seismic datasets. This synthetic approach not only supports neural network training but also allows researchers to investigate advanced supervised Machine Learning techniques with greater flexibility. Consequently, \texttt{PyAWD} stands out as an effective solution for enhancing spatial and temporal pattern recognition studies and offers a compelling alternative to data gathered from physical deployments, as a preliminary study, in cases where this deployment is expensive or difficult.
        Although \texttt{PyAWD} addresses several limitations of traditional seismic data acquisition, it is not without its challenges. One primary limitation is the reliance on synthetic datasets, which, while highly customizable, may not fully capture the complexities of real-world seismic events. Environmental noise, unmodeled material properties, and unexpected wave interactions present in natural data are difficult to simulate accurately. Future work should aim to integrate real-world seismic data with \texttt{PyAWD}-generated synthetic data, to improve model robustness and generalizability. Additionally, exploring the use of more advanced Machine Learning models—such as attention mechanisms and graph-based neural networks—could further enhance the predictive power of models trained on \texttt{PyAWD} data.
        The broader implications of \texttt{PyAWD} go beyond epicenter retrieval. With its ability to simulate a variety of seismic events, including earthquakes and explosions, this tool has the potential to advance multiple geophysical research areas. By enabling control over material properties, wave sources, and propagation fields, researchers can pursue new investigations in fields such as underground resource exploration, earthquake hazard assessment, structural health monitoring, or seismometer placement optimization.
        In conclusion, \texttt{PyAWD} primarily contributes to seismic analysis by providing a highly customizable simulator that enables more accessible and efficient use of Machine Learning. By facilitating the generation of high-quality synthetic datasets, the tool supports the application of supervised learning models in seismic research. Although challenges remain, particularly concerning the integration of synthetic data with real-world scenarios, the potential of \texttt{PyAWD} to streamline seismic dataset generation is promising. Future efforts will focus on expanding the tool’s capabilities, incorporating real-world data, and refining the underlying simulation methods to further enhance its utility in seismic analysis.

\section{Conclusion}
        This paper presents \texttt{PyAWD}, a Python library designed to streamline seismic analysis by combining customizable wave simulations with Machine Learning integration. By enabling synthetic data generation, \texttt{PyAWD} enhances the applicability of supervised learning models in earthquake analysis, supporting hazard assessments and risk mitigation efforts. As seismic events continue to affect communities globally, tools like \texttt{PyAWD} offer new possibilities for informed decision-making and more resilient infrastructure planning.
        Looking forward, several enhancements to \texttt{PyAWD} are anticipated. Simulating other wave equations such as, for example, the elastic wave equation, could extend the scope of the library. The modular design of \texttt{PyAWD} allows for easy adaptation of solver techniques, facilitating experimentation with different equations and initial conditions. Extending the simulator to handle diverse wave sources, such as moving fault segments or time-varying seismic sources, would provide researchers with additional insights into how source variations influence wave propagation. Finally, using transfer learning and domain adaptation, researchers could leverage \texttt{PyAWD} to pre-train models applicable to various geological regions, potentially reducing the need for extensive field measurements in those areas.

\newpage
\textbf{Code availability section}
The author has no competing interests to claim. Pascal Tribel and Gianluca Bontempi are affiliated with \textit{TRusted AI Labs} (TRAIL).
Gianluca Bontempi is supported by the Service Public de Wallonie Recherche under grant nr. 2010235-ARIAC by \textit{DigitalWallonia4.ai}. Computational resources have been provided by the \textit{Consortium des Equipements de Calcul Intensif} (CECI), funded by the \textit{Fonds de la Recherche Scientifique de Belgique} (F.R.S.-FNRS) under Grant No. 2.5020.11 and by the Walloon Region. 
The authors would like to thank Corentin Caudron (ULB) for his insightful remarks.
\begin{itemize}
    \item Name of the Library: PyAWD
    \item Contact: pascal.tribel@ulb.be
    \item Hardware requirements: None
    \item Program language: Python 3.x
\end{itemize}
The source code is available for download at the link: \url{https://github.com/pascaltribel/pyawd}

\bibliographystyle{cas-model2-names}
\bibliography{bibliography} 

\end{document}